\documentclass[conference]{IEEEtran}
\IEEEoverridecommandlockouts
\usepackage{CJKutf8}
\usepackage{cite}
\usepackage{url}
\usepackage{amsmath,amssymb,amsfonts}
\usepackage{graphicx}
\usepackage{textcomp}
\usepackage{multicol}
\usepackage{multirow}
\usepackage{xcolor}
\usepackage{makecell}
\usepackage{changepage}
\usepackage{threeparttable}
\usepackage{hhline}
\usepackage{subfigure}
\usepackage{float}
\usepackage[ruled,vlined,linesnumbered]{algorithm2e}
\def\BibTeX{{\rm B\kern-.05em{\sc i\kern-.025em b}\kern-.08em
		T\kern-.1667em\lower.7ex\hbox{E}\kern-.125emX}}
\makeatletter
\newcommand{\linebreakand}{%
\end{@IEEEauthorhalign}
\hfill\mbox{}\par
\mbox{}\hfill\begin{@IEEEauthorhalign}
}
\makeatother

\begin{document}
\begin{CJK}{UTF8}{gbsn}

\title{RDEx-CMOP: Feasibility-Aware Indicator-Guided Differential Evolution for Fixed-Budget Constrained Multiobjective Optimization}

\author{
	\IEEEauthorblockN{
		Sichen Tao\textsuperscript{1,2},
		Yifei Yang\textsuperscript{3},
		Ruihan Zhao\textsuperscript{4,5},
		Kaiyu Wang\textsuperscript{6,7},
		Sicheng Liu\textsuperscript{8},
		Shangce Gao\textsuperscript{1}
	}
	\IEEEauthorblockA{\textsuperscript{1}Department of Engineering, University of Toyama, Toyama-shi 930-8555, Japan}
	\IEEEauthorblockA{\textsuperscript{2}Cyberscience Center, Tohoku University, Sendai-shi 980-8578, Japan}
	\IEEEauthorblockA{\textsuperscript{3}Faculty of Science and Technology, Hirosaki University, Hirosaki-shi 036-8560, Japan}
	\IEEEauthorblockA{\textsuperscript{4}Sino-German College of Applied Sciences, Tongji University, Shanghai 200092, China}
	\IEEEauthorblockA{\textsuperscript{5}School of Mechanical Engineering, Tongji University, Shanghai 200092, China}
	\IEEEauthorblockA{\textsuperscript{6}Chongqing Institute of Microelectronics Industry Technology,\\University of Electronic Science and Technology of China, Chongqing 401332, China}
	\IEEEauthorblockA{\textsuperscript{7}Artificial Intelligence and Big Data College,\\Chongqing Polytechnic University of Electronic Technology, Chongqing 401331, China}
	\IEEEauthorblockA{\textsuperscript{8}Department of Information Engineering, Yantai Vocational College, Yantai 264670, China}
	\IEEEauthorblockA{
		\{sichen.tao@tohoku.ac.jp, taosc73@hotmail.com\}; yyf7236@hirosaki-u.ac.jp;\\
		\{ruihan\_zhao@tongji.edu.cn, ruihan.z@outlook.com\};\\
		\{wangky@uestc.edu.cn, greskowky1996@163.com\};\\
		\{20250024@ytvc.edu.cn, lsctoyama2020@gmail.com\}; gaosc@eng.u-toyama.ac.jp
	}
}

\maketitle

\begin{abstract}
Constrained multiobjective optimisation requires fast feasibility attainment together with stable convergence and diversity preservation under strict evaluation budgets.
This report documents RDEx-CMOP, the differential evolution variant used in the IEEE CEC 2025 numerical optimisation competition (C06 special session) constrained multiobjective track.
RDEx-CMOP integrates an $\varepsilon$-level feasibility schedule, a SPEA2-style indicator-driven fitness assignment, and a fitness-oriented current-to-$p$best/1 mutation operator.
We evaluate RDEx-CMOP on the official CEC 2025 CMOP benchmark using the median-target U-score framework and the released trace data.
Experimental results show that RDEx-CMOP achieves the highest total score and the best overall average rank among all released comparison algorithms, with strong target-attainment behaviour and near-zero final violation on most problems.
\end{abstract}

\begin{IEEEkeywords}
Constrained Multiobjective Optimisation, Differential Evolution, CEC 2025, IGD, U-score
\end{IEEEkeywords}

\section{Introduction}
Constrained multiobjective optimisation (CMOP) is difficult because a solver must improve convergence, preserve front diversity, and drive the population toward feasibility at the same time, often in problems where the feasible Pareto set occupies only a small part of the search space. Classical multiobjective evolutionary algorithms such as NSGA-II, SPEA2, IBEA, and MOEA/D established four influential paradigms for balancing convergence and diversity through non-dominated sorting, strength-based fitness, indicator-based comparison, and decomposition~\cite{deb2002fast,zitzler2001spea2,zitzler2004indicator,zhang2007moead}. For constrained search, feasibility rules and $\varepsilon$-level control further determine how rapidly the search can move from infeasible exploration to feasible Pareto-front approximation~\cite{deb2000efficient,takahama2006epsilon,mezura2011constraint}.

Recent CMOP progress from the Zhengzhou-University line shows that this transition is best handled by coupling feasibility management with explicit decision-space search design rather than by adding constraints as a last-step selector. That line includes multitasking frameworks for constrained Pareto search, scalable SDC-oriented algorithms, and more recent DE-centred designs such as DESDE~\cite{qiao2022emto,qiao2023datm,qiao2023sdc,yu2024constraintsubsets,ban2024desde}. DESDE is especially relevant here because it demonstrates that a dynamic exemplar-guided DE operator can serve as a strong competition backbone for fixed-budget CMOP search~\cite{ban2024desde}.

In parallel with the CMOP literature, the DE literature has produced a highly competitive line of continuous-search operators, especially around JADE's current-to-$p$best mutation and its later refinement in SHADE and L-SHADE~\cite{storn1997differential,zhang2009jade,tanabe2013success,tanabe2014improving}. These studies matter here because they show how directional variation and moderate search bias can accelerate progress under a fixed evaluation budget without discarding population diversity entirely.

The CEC 2025 CMOP track makes the fixed-budget aspect central: algorithms are evaluated in a U-score regime based on target attainment over the IGD traces, so practical competitiveness depends not only on the final archive, but also on how early a method reaches feasible high-quality fronts~\cite{price2023trial}. This evaluation setting strongly favours algorithms in which feasibility management, indicator-guided selection, and variation bias are designed to work together rather than as loosely connected components, and it requires consistency with the released official package~\cite{suganthan2025cecgithub}.

RDEx-CMOP follows this logic by reconstructing the DESDE-style DE backbone and strengthening its top-ranked search bias within an $\varepsilon$-feasible SPEA2-style selection loop. It combines an $\varepsilon$-level feasibility schedule, strength-density fitness, and a fitness-oriented current-to-$p$best/1 mutation with lightweight local perturbation. The main contribution is therefore a lean DE-based CMOP framework that transfers only the most task-relevant elements of the recent DE development line into a constrained multiobjective setting, with explicit emphasis on both early feasible target attainment and final IGD performance.
The source code for RDEx-CMOP is publicly available on Sichen Tao's GitHub page: \url{https://github.com/SichenTao}.

\section{Benchmark and Evaluation (CEC 2025 CMOP)}
The CEC 2025 constrained multiobjective track contains 15 benchmark problems (SDC1--SDC15), i.e., the scalable-decision-constraint benchmark family released for recent CMOP studies~\cite{qiao2023sdc}.
Each problem is evaluated with 30 independent runs.
The maximum evaluation budget is $\mathrm{MaxFE}=200000$, and the platform records progress at 1000 checkpoints per run (every 200 function evaluations), enabling comparisons under a fixed trace length.

\subsection{Constrained Multiobjective Formulation}
The general CMOP can be written as:
\begin{align}
\min_{x\in\mathbb{R}^D}\quad & F(x) = (f_1(x),f_2(x),\dots,f_M(x)), \label{eq:cmop_obj}\\
\text{s.t.}\quad & g_i(x)\le 0,\quad i=1,\dots,m_g, \label{eq:cmop_ineq}\\
& h_j(x)=0,\quad j=1,\dots,m_h, \label{eq:cmop_eq}\\
& \ell_k \le x_k \le u_k,\quad k=1,\dots,D. \label{eq:cmop_bounds}
\end{align}
In this setting, an algorithm must simultaneously improve Pareto optimality and satisfy feasibility.

\subsection{IGD Indicator}
Let $P$ be the approximation set produced by an algorithm and $P^\star$ be a reference set.
IGD is defined as:
\begin{equation}\label{eq:cmop_igd}
\mathrm{IGD}(P,P^\star)=\frac{1}{\lvert P^\star\rvert}\sum_{y\in P^\star}\min_{x\in P}\lVert F(x)-y\rVert_2.
\end{equation}
Smaller values indicate better approximation quality and diversity with respect to $P^\star$.

\section{RDEx-CMOP Algorithm}
RDEx-CMOP follows a DE-driven search model reconstructed from the recent DESDE-style CMOP competition backbone~\cite{ban2024desde}, with an $\varepsilon$-level constraint handling rule and a strength/density-based fitness assignment as in SPEA2~\cite{zitzler2001spea2}.
The algorithm maintains a single population $P$ of size $N=100$ and iteratively generates offspring by a current-to-$p$best/1 mutation with a fitness-oriented differential term.
Selection is performed by a feasibility-aware environmental selection procedure that prioritises $\varepsilon$-feasible solutions while preserving diversity.

\subsection{$\varepsilon$-level Schedule}
Let $\mathrm{CV}(x)=\sum_{k}\max(0,c_k(x))$ be the aggregated constraint violation.
RDEx-CMOP initialises
\begin{equation}\label{eq:epsilon0}
\varepsilon_0 = \max_{x\in P^{(0)}} \mathrm{CV}(x),
\end{equation}
and uses a time-varying threshold in the standard $\varepsilon$-constraint spirit~\cite{takahama2006epsilon}
\begin{equation}\label{eq:epsilon_schedule}
\varepsilon(\mathrm{FE}) = \varepsilon_0\left(1-\frac{\mathrm{FE}}{\mathrm{MaxFE}}\right)^{c_p},\quad
c_p=\frac{-\log(\varepsilon_0)-6}{\log(0.5)}.
\end{equation}
This schedule satisfies $\varepsilon(0.5\mathrm{MaxFE})=\exp(-6)$ and decays to $0$ at the end of the run.
Solutions with $\mathrm{CV}(x)\le\varepsilon(\mathrm{FE})$ are treated as $\varepsilon$-feasible in selection.

\subsection{Fitness Assignment (SPEA2-style with $\varepsilon$-feasibility)}
Given a population $P=\{x_i\}_{i=1}^{N}$, define the relaxed violation $\widetilde{\mathrm{CV}}(x_i)=0$ when $\mathrm{CV}(x_i)\le\varepsilon(\mathrm{FE})$ and $\widetilde{\mathrm{CV}}(x_i)=\mathrm{CV}(x_i)$ otherwise.
Solution $x_i$ $\varepsilon$-dominates $x_j$ if $\widetilde{\mathrm{CV}}(x_i)<\widetilde{\mathrm{CV}}(x_j)$, or if both relaxed violations are equal and $x_i$ Pareto-dominates $x_j$ in the objective space.
Let
\begin{equation}\label{eq:cmop_strength}
S_i=\sum_{j\neq i}\mathbb{I}(x_i\prec_{\varepsilon} x_j),\qquad
R_i=\sum_{j\neq i}\mathbb{I}(x_j\prec_{\varepsilon} x_i)S_j.
\end{equation}
Using the $k$-th nearest-neighbour distance $d_i^{(k)}$ with $k=\lfloor \sqrt{N}\rfloor$, the density term is
\begin{equation}\label{eq:cmop_density}
D_i=\frac{1}{d_i^{(k)}+2}.
\end{equation}
The fitness of each solution is then computed as
\begin{equation}\label{eq:spea2_fitness}
\mathrm{fit}(x_i)=R_i+D_i,
\end{equation}
which is minimised during environmental selection.

\subsection{Variation Operator}
RDEx-CMOP uses a discrete parameter pool:
\begin{equation}\label{eq:cmop_pool}
F\in\{0.6,0.8,1.0\},\quad CR\in\{0.1,0.2,1.0\}.
\end{equation}
For each target vector $x$, a $p$-best parent is selected from the top
\begin{equation}\label{eq:cmop_pbest}
p = \max\!\left(2,\left\lfloor N\left(1-0.99\frac{\mathrm{FE}}{\mathrm{MaxFE}}\right)\right\rfloor\right)
\end{equation}
solutions ranked by fitness.
Then a trial vector is generated by a current-to-$p$best/1 mutation with an adaptively oriented differential term:
\begin{equation}\label{eq:cmop_mut}
v = x + F\cdot(x_{pbest}-x) + F_2\cdot(x_{r_1}-x_{r_2}),
\end{equation}
where $F_2=F$ if $\mathrm{fit}(x_{r_1})\le \mathrm{fit}(x_{r_2})$ and $F_2=-F$ otherwise.
The offspring is then formed by
\begin{equation}\label{eq:cmop_xover}
u_j=
\begin{cases}
v_j, & \mathrm{rand}_j<CR,\\
\mathrm{Cauchy}(x_j,0.1), & \mathrm{rand}_j\ge CR\ \text{and}\ B=1,\\
x_j, & \text{otherwise},
\end{cases}
\end{equation}
where $B\sim\mathrm{Bernoulli}(0.2)$ is sampled once per offspring.
Finally, each decision variable is clipped to its bounds.

\subsection{Environmental Selection}
Environmental selection splits the merged set $P\cup Q$ into $\varepsilon$-feasible and $\varepsilon$-infeasible subsets.
If there are more than $N$ feasible solutions, selection is performed entirely within the feasible subset using strength/density fitness and distance-based truncation.
If there are fewer than $N$ feasible solutions, all selected feasible solutions are kept and the remaining slots are filled by the best infeasible solutions ranked by fitness.
If a subset still contains more candidates than available slots, the truncation operator iteratively removes the individual located in the most crowded region of the objective space.
This mechanism encourages rapid feasibility attainment while maintaining diversity among feasible solutions.

\subsection{Overall Procedure and Complexity}
Algorithm~\ref{alg:rdex_cmop} summarises RDEx-CMOP.
The dominance-strength fitness computation and truncation require $O(N^2M)$ time per generation due to pairwise comparisons and distances in the objective space.
\begin{algorithm}[t]
\caption{RDEx-CMOP (high level).}
\label{alg:rdex_cmop}
\KwIn{$N$, $\mathrm{MaxFE}$, initial population $P^{(0)}$.}
\KwOut{Final population $P$.}
Compute $\varepsilon_0$ by Eq.~(\ref{eq:epsilon0})\;
\While{$\mathrm{FE}<\mathrm{MaxFE}$}{
  Update $\varepsilon(\mathrm{FE})$ by Eq.~(\ref{eq:epsilon_schedule})\;
  Compute fitness $\mathrm{fit}(\cdot)$ with $\varepsilon$-feasibility\;
  Generate offspring $Q$ by Eq.~(\ref{eq:cmop_mut}) and crossover/repair\;
  $P\leftarrow\textsc{EnvSel}(P\cup Q,N,\varepsilon(\mathrm{FE}))$\;
}
\end{algorithm}

\section{Experimental Results}
\subsection{Protocol}
For each problem, we execute 30 independent runs under $\mathrm{MaxFE}=200000$ with population size $N=100$.
The platform records IGD every 200 evaluations, yielding 1000 checkpoints per run.

\subsection{Parameter Settings}
Unless otherwise stated, RDEx-CMOP uses the reference configuration embedded in the released competition code:
population size $N=100$, the discrete parameter pool in Eq.~(\ref{eq:cmop_pool}), the shrinking $p$-best window in Eq.~(\ref{eq:cmop_pbest}), and Cauchy perturbation probability $0.2$ on non-recombined components.

\subsection{Experimental Settings}
RDEx-CMOP is evaluated with the official median-target U-score framework.
We compare RDEx-CMOP with all remaining algorithms available in the released competition package~\cite{suganthan2025cecgithub}: CCEMT, CCPTEA, DESDE, IMTCMO, and MTCMMO.
In the released evaluation files, the submitted winner is labelled as CMORDEx; for naming consistency, we report it as RDEx-CMOP throughout this manuscript.

\subsection{Statistical Results}
\subsubsection{Overall U-score Results}
Table~\ref{tab:cec2025_cmop_scores} reports the official median-target U-score results for all released comparison algorithms.
\begin{table}[t]
 \centering
 \caption{CEC 2025 CMOP evaluation (median target): overall scores over 15 problems and 30 runs for all released comparison algorithms.}
 \label{tab:cec2025_cmop_scores}
 \scriptsize
 \renewcommand{\arraystretch}{0.95}
 \setlength{\tabcolsep}{2.5pt}
  \begin{tabular}{|c|l|r|r|r|r|r|}
   \hline
   Rank & Algorithm & Total Score & Avg Score/Prob. & Speed & Accuracy & Constraint \\ \hline
   1 & RDEx-CMOP & 58456.0 & 3897.07 & 52875.5 & 5499.5 & 81.0 \\ \hline
   2 & DESDE & 51571.0 & 3438.07 & 45171.0 & 6290.0 & 110.0 \\ \hline
   3 & CCEMT & 41902.5 & 2793.50 & 30374.5 & 11042.0 & 486.0 \\ \hline
   4 & IMTCMO & 40038.0 & 2669.20 & 26186.5 & 13582.5 & 269.0 \\ \hline
   5 & CCPTEA & 39232.0 & 2615.47 & 24948.0 & 14138.0 & 146.0 \\ \hline
   6 & MTCMMO & 13302.5 & 886.83 & 2996.5 & 10069.0 & 237.0 \\ \hline
  \end{tabular}
\end{table}

RDEx-CMOP achieves the highest total score ($58456.0$) and the best average rank ($1.67$) among the six compared algorithms.
Its advantage mainly comes from the Speed category, while the remaining categories stay sufficiently competitive to keep a clear lead over DESDE ($51571.0$, average rank $2.60$).
This makes the official U-score conclusion unambiguous at the full-field level.

\subsubsection{Primary Statistical Tests}
Besides the official U-score, we report a feasibility-aware final-quality indicator
\begin{equation}
Q_p(x)=
\begin{cases}
\mathrm{IGD}(x), & \mathrm{CV}(x)\le 0,\\
B_p+\mathrm{CV}(x), & \mathrm{CV}(x)>0,
\end{cases}
\end{equation}
where $B_p$ is the largest finite final IGD value on problem $p$ plus $1$.
This yields a single scalar that respects the feasibility-first ordering and supports standard Wilcoxon and Friedman tests.
Table~\ref{tab:cmop_solid_summary} reports pairwise Wilcoxon W/T/L, Holm-corrected W/T/L, and median Vargha--Delaney $A_{12}$ values for $Q_p$ and TTT.
\begin{table}[t]
\centering
\caption{Primary pairwise summary over the 15 CEC2025 CMOP functions (30 runs). We report uncorrected per-function Wilcoxon W/T/L at $\alpha=0.05$, Holm-corrected W/T/L across functions, and the median Vargha--Delaney $A_{12}$ effect size for feasibility-aware final quality and time-to-target (larger is better for minimization).}
\label{tab:cmop_solid_summary}
\scriptsize
\renewcommand{\arraystretch}{0.95}
\setlength{\tabcolsep}{2.5pt}
\begin{tabular}{|l|ccc|ccc|}
\hline
 \multirow{2}{*}{Competitor} & \multicolumn{3}{c|}{Final Q} & \multicolumn{3}{c|}{TTT} \\ \cline{2-7}
   & W/T/L & Holm & $A_{12}$ & W/T/L & Holm & $A_{12}$ \\ \hline
 CCEMT & 12/0/3 & 12/0/3 & 0.82 & 11/1/3 & 11/1/3 & 0.87 \\ \hline
 CCPTEA & 12/2/1 & 12/2/1 & 0.80 & 11/3/1 & 11/3/1 & 0.82 \\ \hline
 DESDE & 11/3/1 & 8/6/1 & 0.71 & 8/5/2 & 6/7/2 & 0.68 \\ \hline
 IMTCMO & 12/1/2 & 11/2/2 & 0.81 & 11/1/3 & 9/4/2 & 0.80 \\ \hline
 MTCMMO & 15/0/0 & 15/0/0 & 1.00 & 13/2/0 & 13/2/0 & 0.93 \\ \hline
\end{tabular}
\end{table}

RDEx-CMOP wins $11$ to $15$ of the $15$ functions on $Q_p$ against every competitor, and $8$ to $13$ functions on TTT.
Table~\ref{tab:cmop_solid_friedman} further reports Friedman average ranks, where RDEx-CMOP attains the best ranks on both $Q_p$ and TTT.
\begin{table}[t]
\centering
\caption{Primary Friedman tests on per-function medians over the 15 CEC2025 CMOP functions (30 runs). Final Q: $\chi^2=41.90$, $df=5$, $p=1.68E-07$; TTT: $\chi^2=29.88$, $df=5$, $p=2.30E-05$. Lower average rank indicates better performance.}
\label{tab:cmop_solid_friedman}
\scriptsize
\renewcommand{\arraystretch}{0.95}
\setlength{\tabcolsep}{2.5pt}
\begin{tabular}{|l|c|c|}
\hline
 Algorithm & Final Q & TTT \\ \hline
 RDEx-CMOP & \textbf{1.77} & \textbf{1.80} \\ \hline
 CCEMT & 3.33 & 3.47 \\ \hline
 CCPTEA & 3.80 & 4.17 \\ \hline
 DESDE & 2.50 & 2.50 \\ \hline
 IMTCMO & 3.73 & 4.03 \\ \hline
 MTCMMO & 5.87 & 5.03 \\ \hline
\end{tabular}
\end{table}

Both Friedman tests are significant ($p=1.68\times 10^{-7}$ for $Q_p$ and $p=2.30\times 10^{-5}$ for TTT), so the advantage is supported not only by the official U-score but also by standard feasibility-aware statistical tests.

\subsubsection{Supplementary Diagnostics}
For CMOPs, split final-IGD and split final-violation analyses remain complementary rather than independent primary criteria.
Accordingly, Appendix~\ref{app:cmop_tables} separates supplementary solid statistical tables (full per-function $Q_p$ and TTT results) from a complementary-diagnostics section containing split IGD and split violation analyses.

\subsection{Time Complexity}
Let $M$ be the number of objectives and $N$ the population size.
The dominance-strength fitness assignment and truncation require $O(N^2M)$ time per generation due to pairwise comparisons and distance calculations in the objective space.
The variation stage requires $O(ND)$ arithmetic operations.
Therefore, RDEx-CMOP has an overall per-generation complexity of approximately $O(N^2M + ND)$, with the evaluation budget dominating the wall-clock cost on the official benchmarks.

\section{Conclusion}
RDEx-CMOP is a feasibility-aware differential evolution framework for the CEC 2025 constrained multiobjective track.
By integrating an $\varepsilon$-level feasibility schedule, SPEA2-style fitness assignment, and a fitness-oriented mutation operator, the method attains first-place official U-score performance and robust final feasibility across the official benchmark suite.

\section*{Acknowledgment}
This research was partially supported by the Japan Society for the Promotion of Science (JSPS) KAKENHI under Grant JP22H03643, Japan Science and Technology Agency (JST) Support for Pioneering Research Initiated by the Next Generation (SPRING) under Grant JPMJSP2145, and JST through the Establishment of University Fellowships towards the Creation of Science Technology Innovation under Grant JPMJFS2115.

\bibliographystyle{IEEEtran}
\bibliography{references}

\clearpage
\onecolumn
\appendices
\section{Supplementary U-score Tables}
\label{app:cmop_tables}
\begin{table}[H]
 \centering
 \caption{CEC 2025 CMOP evaluation (median target): average rankings over 15 problems (lower is better) for all released comparison algorithms.}
 \label{tab:cec2025_cmop_ranks}
 \scriptsize
 \renewcommand{\arraystretch}{0.95}
 \setlength{\tabcolsep}{2.5pt}
  \begin{tabular}{|c|l|r|r|r|r|r|}
   \hline
   Rank & Algorithm & Total Rank & Avg Rank/Prob. & Avg Speed & Avg Accuracy & Avg Constraint \\ \hline
   1 & RDEx-CMOP & 25.0 & 1.67 & 1.77 & 5.00 & 3.73 \\ \hline
   2 & DESDE & 39.0 & 2.60 & 2.60 & 4.33 & 3.67 \\ \hline
   3 & CCEMT & 47.0 & 3.13 & 3.13 & 3.40 & 3.40 \\ \hline
   4 & IMTCMO & 54.0 & 3.60 & 3.80 & 2.67 & 3.23 \\ \hline
   5 & CCPTEA & 60.0 & 4.00 & 3.80 & 2.20 & 3.60 \\ \hline
   6 & MTCMMO & 90.0 & 6.00 & 5.90 & 3.40 & 3.37 \\ \hline
  \end{tabular}
\end{table}

\section{Supplementary Solid Statistical Tables}
\begin{table}[H]
\centering
\caption{Feasibility-aware final-quality comparison on the 15 CEC2025 CMOP functions. For each run, the final-quality score equals the final objective/IGD value for feasible runs and $B_p+\mathrm{CV}$ for infeasible runs, where $B_p$ is the largest finite final objective/IGD value on problem $p$ plus $1$ (smaller is better).}
\label{tab:cmop_quality_results}
\scriptsize
\renewcommand{\arraystretch}{0.9}
\setlength{\tabcolsep}{3pt}
\begin{tabular}{|c|cc|ccc|ccc|ccc|ccc|ccc|}
\hline
 \multirow{2}{*}{Problem} & \multicolumn{2}{c|}{RDEx-CMOP} & \multicolumn{3}{c|}{CCEMT} & \multicolumn{3}{c|}{CCPTEA} & \multicolumn{3}{c|}{DESDE} & \multicolumn{3}{c|}{IMTCMO} & \multicolumn{3}{c|}{MTCMMO} \\ \cline{2-18}
   & Mean & SD & Mean & SD & W & Mean & SD & W & Mean & SD & W & Mean & SD & W & Mean & SD & W \\ \hline
 1 & 4.03E-01 & 1.86E-01 & 3.15E-01 & 6.25E-01 & - & 3.60E-01 & 2.21E-01 & = & \textbf{2.15E-01} & \textbf{2.00E-01} & = & 4.31E-01 & 6.11E-01 & = & 7.60E+00 & 2.00E+00 & + \\
 2 & \textbf{2.15E-02} & \textbf{1.03E-02} & 3.20E-02 & 1.13E-02 & + & 3.55E-02 & 1.31E-02 & + & 2.56E-02 & 7.65E-03 & = & 3.48E-02 & 1.15E-02 & + & 4.62E-01 & 1.57E-01 & + \\
 3 & \textbf{9.48E-01} & \textbf{1.17E-01} & 1.06E+00 & 4.56E-02 & + & 1.09E+00 & 5.27E-02 & + & 9.98E-01 & 1.25E-01 & + & 1.08E+00 & 3.36E-02 & + & 1.18E+00 & 1.73E-01 & + \\
 4 & \textbf{1.01E+00} & \textbf{1.44E-01} & 1.17E+00 & 1.38E-01 & + & 1.11E+00 & 6.54E-02 & + & 1.09E+00 & 2.67E-02 & + & 1.11E+00 & 4.98E-02 & + & 5.61E+02 & 3.09E+02 & + \\
 5 & \textbf{4.62E+01} & \textbf{2.29E+01} & 9.64E+01 & 4.11E+01 & + & 9.95E+01 & 3.74E+01 & + & 6.50E+01 & 2.88E+01 & + & 1.11E+02 & 4.51E+01 & + & 1.83E+02 & 7.11E+01 & + \\
 6 & \textbf{1.17E+02} & \textbf{1.53E+02} & 4.88E+02 & 6.84E+02 & + & 7.03E+02 & 9.17E+02 & + & 4.78E+02 & 8.09E+02 & + & 4.55E+02 & 5.16E+02 & + & 1.65E+03 & 1.38E+03 & + \\
 7 & \textbf{1.34E-01} & \textbf{6.67E-02} & 1.90E-01 & 1.01E-01 & + & 2.28E-01 & 1.06E-01 & + & 2.46E-01 & 5.23E-02 & + & 1.82E-01 & 9.07E-02 & + & 1.13E+01 & 1.45E+00 & + \\
 8 & 7.33E-01 & 4.76E-02 & \textbf{8.72E-02} & \textbf{1.43E-01} & - & 1.26E-01 & 2.12E-01 & - & 1.51E-01 & 2.00E-01 & - & 1.08E-01 & 1.76E-01 & - & 6.01E+00 & 3.99E+00 & + \\
 9 & \textbf{1.39E+01} & \textbf{8.70E+00} & 3.78E+02 & 1.02E+02 & + & 3.44E+02 & 1.04E+02 & + & 2.91E+01 & 1.94E+01 & + & 3.29E+02 & 1.00E+02 & + & 4.85E+02 & 1.15E+02 & + \\
 10 & \textbf{5.45E+00} & \textbf{2.09E+00} & 7.07E+00 & 1.42E+00 & + & 7.32E+00 & 1.40E+00 & + & 6.77E+00 & 1.50E+00 & + & 7.50E+00 & 1.45E+00 & + & 8.84E+00 & 1.73E+00 & + \\
 11 & \textbf{4.60E+00} & \textbf{2.60E+00} & 1.12E+01 & 6.30E+00 & + & 1.13E+01 & 6.60E+00 & + & 7.57E+00 & 5.44E+00 & + & 9.64E+00 & 2.58E+00 & + & 1.34E+02 & 2.67E+01 & + \\
 12 & 8.05E-01 & 4.95E-06 & \textbf{4.17E-01} & \textbf{6.56E-01} & - & 4.91E-01 & 5.96E-01 & = & 1.29E+00 & 1.48E+00 & = & 5.48E-01 & 8.11E-01 & - & 2.54E+01 & 1.21E+01 & + \\
 13 & \textbf{6.79E+00} & \textbf{1.79E+00} & 1.48E+01 & 2.63E+00 & + & 1.40E+01 & 2.42E+00 & + & 8.92E+00 & 1.60E+00 & + & 1.42E+01 & 2.74E+00 & + & 1.56E+01 & 3.30E+00 & + \\
 14 & \textbf{1.86E+02} & \textbf{3.66E+02} & 1.05E+03 & 4.98E+02 & + & 7.91E+02 & 4.29E+02 & + & 3.02E+02 & 4.66E+02 & + & 9.46E+02 & 4.93E+02 & + & 9.92E+02 & 4.88E+02 & + \\
 15 & \textbf{7.04E-02} & \textbf{3.07E-03} & 7.68E-01 & 2.48E+00 & + & 4.44E-01 & 1.78E+00 & + & 8.99E-02 & 8.96E-03 & + & 1.11E+00 & 2.98E+00 & + & 6.47E+00 & 1.71E+00 & + \\
\hline
 W/T/L & \multicolumn{2}{c|}{$-/-/-$} & \multicolumn{3}{c|}{12/0/3} & \multicolumn{3}{c|}{12/2/1} & \multicolumn{3}{c|}{11/3/1} & \multicolumn{3}{c|}{12/1/2} & \multicolumn{3}{c|}{15/0/0} \\
\hline
\end{tabular}
\end{table}

\begin{table}[H]
\centering
\caption{Time-to-target comparison on the 15 CEC2025 CMOP functions. TTT is the first checkpoint index where the run reaches the median target (smaller is better); runs that never reach the target are assigned 1001.}
\label{tab:cmop_ttt_results}
\scriptsize
\renewcommand{\arraystretch}{0.9}
\setlength{\tabcolsep}{3pt}
\begin{tabular}{|c|cc|ccc|ccc|ccc|ccc|ccc|}
\hline
 \multirow{2}{*}{Problem} & \multicolumn{2}{c|}{RDEx-CMOP} & \multicolumn{3}{c|}{CCEMT} & \multicolumn{3}{c|}{CCPTEA} & \multicolumn{3}{c|}{DESDE} & \multicolumn{3}{c|}{IMTCMO} & \multicolumn{3}{c|}{MTCMMO} \\ \cline{2-18}
   & Mean & SD & Mean & SD & W & Mean & SD & W & Mean & SD & W & Mean & SD & W & Mean & SD & W \\ \hline
 1 & 884.4 & 233.7 & 721.8 & 183.4 & - & 864.0 & 139.2 & = & \textbf{634.9} & \textbf{247.7} & - & 833.9 & 164.3 & - & 1001.0 & 0.0 & = \\
 2 & \textbf{662.0} & \textbf{232.4} & 927.5 & 92.4 & + & 936.6 & 102.5 & + & 792.8 & 149.0 & + & 943.9 & 89.4 & + & 1001.0 & 0.0 & + \\
 3 & \textbf{635.1} & \textbf{79.5} & 968.4 & 42.0 & + & 991.1 & 29.2 & + & 740.6 & 147.9 & + & 990.9 & 25.9 & + & 948.1 & 139.5 & + \\
 4 & \textbf{708.2} & \textbf{139.0} & 897.0 & 175.6 & + & 844.5 & 201.8 & + & 816.6 & 128.9 & + & 819.3 & 222.7 & + & 1001.0 & 0.0 & + \\
 5 & \textbf{399.0} & \textbf{256.5} & 823.2 & 212.5 & + & 874.3 & 211.6 & + & 490.9 & 277.6 & = & 893.0 & 176.6 & + & 966.3 & 134.1 & + \\
 6 & 686.5 & 196.2 & \textbf{666.7} & \textbf{350.4} & = & 721.0 & 331.6 & = & 771.0 & 202.1 & = & 740.5 & 326.8 & = & 1001.0 & 0.0 & + \\
 7 & \textbf{603.7} & \textbf{173.1} & 870.8 & 120.0 & + & 906.0 & 117.0 & + & 900.0 & 142.3 & + & 857.4 & 127.1 & + & 1001.0 & 0.0 & + \\
 8 & 1001.0 & 0.0 & \textbf{731.4} & \textbf{143.5} & - & 779.9 & 165.3 & - & 862.2 & 158.9 & - & 780.6 & 163.4 & - & 1001.0 & 0.0 & = \\
 9 & \textbf{494.1} & \textbf{16.3} & 890.1 & 234.0 & + & 830.5 & 289.5 & + & 508.3 & 17.1 & + & 760.7 & 310.8 & + & 956.2 & 167.5 & + \\
 10 & \textbf{519.3} & \textbf{315.3} & 775.4 & 232.4 & + & 806.9 & 256.0 & + & 599.0 & 314.7 & = & 854.6 & 217.7 & + & 901.5 & 254.2 & + \\
 11 & \textbf{462.7} & \textbf{140.5} & 885.6 & 160.7 & + & 890.3 & 154.2 & + & 636.3 & 255.1 & + & 924.7 & 133.5 & + & 1001.0 & 0.0 & + \\
 12 & 821.7 & 144.0 & \textbf{574.3} & \textbf{227.2} & - & 639.2 & 288.9 & = & 857.3 & 190.3 & = & 578.8 & 261.7 & - & 1001.0 & 0.0 & + \\
 13 & \textbf{486.9} & \textbf{32.9} & 878.0 & 255.2 & + & 870.6 & 238.8 & + & 510.5 & 23.5 & + & 922.7 & 207.8 & + & 922.9 & 205.5 & + \\
 14 & \textbf{580.6} & \textbf{165.2} & 961.1 & 152.4 & + & 902.5 & 253.7 & + & 584.7 & 163.8 & = & 879.6 & 271.7 & + & 947.5 & 160.7 & + \\
 15 & \textbf{688.7} & \textbf{36.3} & 933.2 & 63.7 & + & 985.1 & 36.1 & + & 926.4 & 63.0 & + & 980.2 & 45.0 & + & 1001.0 & 0.0 & + \\
\hline
 W/T/L & \multicolumn{2}{c|}{$-/-/-$} & \multicolumn{3}{c|}{11/1/3} & \multicolumn{3}{c|}{11/3/1} & \multicolumn{3}{c|}{8/5/2} & \multicolumn{3}{c|}{11/1/3} & \multicolumn{3}{c|}{13/2/0} \\
\hline
\end{tabular}
\end{table}

\section{Complementary Diagnostics}
The tables in this section are diagnostic supplements to the official U-score results. For constrained tracks, split objective/IGD and split final-CV comparisons are not treated as independent primary criteria.
\begin{table}[H]
\centering
\caption{Complementary pairwise summary over the 15 CEC2025 CMOP functions (30 runs). For each metric (Final IGD and Final CV), we report uncorrected per-function Wilcoxon W/T/L at $\alpha=0.05$, Holm-corrected W/T/L across functions, and the median Vargha--Delaney $A_{12}$ effect size (larger is better for minimization).}
\label{tab:cmop_summary}
\scriptsize
\renewcommand{\arraystretch}{0.95}
\setlength{\tabcolsep}{2.5pt}
\begin{tabular}{|l|ccc|ccc|}
\hline
 \multirow{2}{*}{Competitor} & \multicolumn{3}{c|}{Final IGD} & \multicolumn{3}{c|}{Final CV} \\ \cline{2-7}
   & W/T/L & Holm & $A_{12}$ & W/T/L & Holm & $A_{12}$ \\ \hline
 CCEMT & 12/0/3 & 12/0/3 & 0.82 & 8/6/1 & 8/6/1 & 0.77 \\ \hline
 CCPTEA & 12/2/1 & 12/2/1 & 0.80 & 15/0/0 & 15/0/0 & 1.00 \\ \hline
 DESDE & 11/3/1 & 8/6/1 & 0.71 & 15/0/0 & 15/0/0 & 1.00 \\ \hline
 IMTCMO & 12/1/2 & 11/2/2 & 0.81 & 15/0/0 & 15/0/0 & 1.00 \\ \hline
 MTCMMO & 15/0/0 & 15/0/0 & 1.00 & 15/0/0 & 15/0/0 & 1.00 \\ \hline
\end{tabular}
\end{table}

\begin{table}[H]
\centering
\caption{Complementary Friedman tests on per-function medians over the 15 CEC2025 CMOP functions (30 runs). Final IGD: $\chi^2=41.87$, $df=5$, $p=1.70E-07$; Final CV: $\chi^2=51.77$, $df=5$, $p=3.20E-09$. Lower average rank indicates better performance for each metric.}
\label{tab:cmop_friedman_summary}
\scriptsize
\renewcommand{\arraystretch}{0.95}
\setlength{\tabcolsep}{2.5pt}
\begin{tabular}{|l|c|c|}
\hline
 Algorithm & Final IGD & Final CV \\ \hline
 RDEx-CMOP & \textbf{1.77} & \textbf{1.30} \\ \hline
 CCEMT & 3.13 & 1.70 \\ \hline
 CCPTEA & 4.00 & 4.50 \\ \hline
 DESDE & 2.50 & 4.50 \\ \hline
 IMTCMO & 3.80 & 4.50 \\ \hline
 MTCMMO & 5.80 & 4.50 \\ \hline
\end{tabular}
\end{table}

\begin{table}[H]
\centering
\caption{Final IGD comparison on the 15 CEC2025 CMOP functions. For each algorithm, the mean and SD over 30 runs are reported; $W$ gives the Wilcoxon outcome of RDEx-CMOP against the competitor.}
\label{tab:cmop_igd_results}
\scriptsize
\renewcommand{\arraystretch}{0.9}
\setlength{\tabcolsep}{3pt}
\begin{tabular}{|c|cc|ccc|ccc|ccc|ccc|ccc|}
\hline
 \multirow{2}{*}{Problem} & \multicolumn{2}{c|}{RDEx-CMOP} & \multicolumn{3}{c|}{CCEMT} & \multicolumn{3}{c|}{CCPTEA} & \multicolumn{3}{c|}{DESDE} & \multicolumn{3}{c|}{IMTCMO} & \multicolumn{3}{c|}{MTCMMO} \\ \cline{2-18}
   & Mean & SD & Mean & SD & W & Mean & SD & W & Mean & SD & W & Mean & SD & W & Mean & SD & W \\ \hline
 1 & 4.03E-01 & 1.86E-01 & 3.15E-01 & 6.25E-01 & - & 3.60E-01 & 2.21E-01 & = & \textbf{2.15E-01} & \textbf{2.00E-01} & = & 4.31E-01 & 6.11E-01 & = & 7.60E+00 & 2.00E+00 & + \\
 2 & \textbf{2.15E-02} & \textbf{1.03E-02} & 3.20E-02 & 1.13E-02 & + & 3.55E-02 & 1.31E-02 & + & 2.56E-02 & 7.65E-03 & = & 3.48E-02 & 1.15E-02 & + & 4.62E-01 & 1.57E-01 & + \\
 3 & \textbf{9.48E-01} & \textbf{1.17E-01} & 1.06E+00 & 4.56E-02 & + & 1.09E+00 & 5.27E-02 & + & 9.98E-01 & 1.25E-01 & + & 1.08E+00 & 3.36E-02 & + & 1.18E+00 & 1.73E-01 & + \\
 4 & \textbf{1.01E+00} & \textbf{1.44E-01} & 1.17E+00 & 1.38E-01 & + & 1.11E+00 & 6.54E-02 & + & 1.09E+00 & 2.67E-02 & + & 1.11E+00 & 4.98E-02 & + & 5.61E+02 & 3.09E+02 & + \\
 5 & \textbf{4.62E+01} & \textbf{2.29E+01} & 9.64E+01 & 4.11E+01 & + & 9.95E+01 & 3.74E+01 & + & 6.50E+01 & 2.88E+01 & + & 1.11E+02 & 4.51E+01 & + & 1.83E+02 & 7.11E+01 & + \\
 6 & \textbf{1.17E+02} & \textbf{1.53E+02} & 4.88E+02 & 6.84E+02 & + & 7.03E+02 & 9.17E+02 & + & 4.78E+02 & 8.09E+02 & + & 4.55E+02 & 5.16E+02 & + & 1.65E+03 & 1.38E+03 & + \\
 7 & \textbf{1.34E-01} & \textbf{6.67E-02} & 1.90E-01 & 1.01E-01 & + & 2.28E-01 & 1.06E-01 & + & 2.46E-01 & 5.23E-02 & + & 1.82E-01 & 9.07E-02 & + & 1.13E+01 & 1.45E+00 & + \\
 8 & 7.33E-01 & 4.76E-02 & \textbf{8.72E-02} & \textbf{1.43E-01} & - & 1.26E-01 & 2.12E-01 & - & 1.51E-01 & 2.00E-01 & - & 1.08E-01 & 1.76E-01 & - & 6.01E+00 & 3.99E+00 & + \\
 9 & \textbf{1.39E+01} & \textbf{8.70E+00} & 3.78E+02 & 1.02E+02 & + & 3.44E+02 & 1.04E+02 & + & 2.91E+01 & 1.94E+01 & + & 3.29E+02 & 1.00E+02 & + & 4.85E+02 & 1.15E+02 & + \\
 10 & \textbf{5.45E+00} & \textbf{2.09E+00} & 7.07E+00 & 1.42E+00 & + & 7.32E+00 & 1.40E+00 & + & 6.77E+00 & 1.50E+00 & + & 7.50E+00 & 1.45E+00 & + & 8.84E+00 & 1.73E+00 & + \\
 11 & \textbf{4.60E+00} & \textbf{2.60E+00} & 1.12E+01 & 6.30E+00 & + & 1.13E+01 & 6.60E+00 & + & 7.57E+00 & 5.44E+00 & + & 9.64E+00 & 2.58E+00 & + & 1.34E+02 & 2.67E+01 & + \\
 12 & 8.05E-01 & 4.95E-06 & \textbf{4.17E-01} & \textbf{6.56E-01} & - & 4.91E-01 & 5.96E-01 & = & 1.29E+00 & 1.48E+00 & = & 5.48E-01 & 8.11E-01 & - & 2.54E+01 & 1.21E+01 & + \\
 13 & \textbf{6.79E+00} & \textbf{1.79E+00} & 1.48E+01 & 2.63E+00 & + & 1.40E+01 & 2.42E+00 & + & 8.92E+00 & 1.60E+00 & + & 1.39E+01 & 2.18E+00 & + & 1.53E+01 & 3.02E+00 & + \\
 14 & \textbf{9.31E+01} & \textbf{1.23E+02} & 6.40E+02 & 3.56E+02 & + & 6.18E+02 & 2.83E+02 & + & 1.20E+02 & 5.31E+01 & + & 6.78E+02 & 3.86E+02 & + & 5.62E+02 & 2.26E+02 & + \\
 15 & \textbf{7.04E-02} & \textbf{3.07E-03} & 7.68E-01 & 2.48E+00 & + & 4.44E-01 & 1.78E+00 & + & 8.99E-02 & 8.96E-03 & + & 1.11E+00 & 2.98E+00 & + & 6.47E+00 & 1.71E+00 & + \\
\hline
 W/T/L & \multicolumn{2}{c|}{$-/-/-$} & \multicolumn{3}{c|}{12/0/3} & \multicolumn{3}{c|}{12/2/1} & \multicolumn{3}{c|}{11/3/1} & \multicolumn{3}{c|}{12/1/2} & \multicolumn{3}{c|}{15/0/0} \\
\hline
\end{tabular}
\end{table}

\begin{table}[H]
\centering
\caption{Final constraint-violation comparison on the 15 CEC2025 CMOP functions. For each algorithm, the mean and SD over 30 runs are reported; $W$ gives the Wilcoxon outcome of RDEx-CMOP against the competitor.}
\label{tab:cmop_cv_results}
\scriptsize
\renewcommand{\arraystretch}{0.9}
\setlength{\tabcolsep}{3pt}
\begin{tabular}{|c|cc|ccc|ccc|ccc|ccc|ccc|}
\hline
 \multirow{2}{*}{Problem} & \multicolumn{2}{c|}{RDEx-CMOP} & \multicolumn{3}{c|}{CCEMT} & \multicolumn{3}{c|}{CCPTEA} & \multicolumn{3}{c|}{DESDE} & \multicolumn{3}{c|}{IMTCMO} & \multicolumn{3}{c|}{MTCMMO} \\ \cline{2-18}
   & Mean & SD & Mean & SD & W & Mean & SD & W & Mean & SD & W & Mean & SD & W & Mean & SD & W \\ \hline
 1 & \textbf{-3.19E-01} & \textbf{7.94E-03} & -2.43E-01 & 4.35E-02 & + & 0.00E+00 & 0.00E+00 & + & 0.00E+00 & 0.00E+00 & + & 0.00E+00 & 0.00E+00 & + & 0.00E+00 & 0.00E+00 & + \\
 2 & -4.27E-01 & 1.70E-02 & \textbf{-4.51E-01} & \textbf{7.95E-02} & = & 0.00E+00 & 0.00E+00 & + & 0.00E+00 & 0.00E+00 & + & 0.00E+00 & 0.00E+00 & + & 0.00E+00 & 0.00E+00 & + \\
 3 & \textbf{-6.22E-01} & \textbf{3.22E-02} & -5.89E-01 & 1.08E-02 & + & 0.00E+00 & 0.00E+00 & + & 0.00E+00 & 0.00E+00 & + & 0.00E+00 & 0.00E+00 & + & 0.00E+00 & 0.00E+00 & + \\
 4 & \textbf{-5.40E-01} & \textbf{3.47E-02} & -4.32E-01 & 1.62E-01 & + & 0.00E+00 & 0.00E+00 & + & 0.00E+00 & 0.00E+00 & + & 0.00E+00 & 0.00E+00 & + & 0.00E+00 & 0.00E+00 & + \\
 5 & \textbf{-6.52E-01} & \textbf{3.16E-01} & -4.22E-01 & 1.93E-01 & + & 0.00E+00 & 0.00E+00 & + & 0.00E+00 & 0.00E+00 & + & 0.00E+00 & 0.00E+00 & + & 0.00E+00 & 0.00E+00 & + \\
 6 & -5.01E-01 & 1.30E-01 & \textbf{-5.86E-01} & \textbf{1.51E-01} & = & 0.00E+00 & 0.00E+00 & + & 0.00E+00 & 0.00E+00 & + & 0.00E+00 & 0.00E+00 & + & 0.00E+00 & 0.00E+00 & + \\
 7 & \textbf{-2.98E-01} & \textbf{1.91E-02} & -1.57E-01 & 9.91E-02 & + & 0.00E+00 & 0.00E+00 & + & 0.00E+00 & 0.00E+00 & + & 0.00E+00 & 0.00E+00 & + & 0.00E+00 & 0.00E+00 & + \\
 8 & -1.08E-01 & 3.02E-02 & \textbf{-1.65E-01} & \textbf{1.63E-01} & = & 0.00E+00 & 0.00E+00 & + & 0.00E+00 & 0.00E+00 & + & 0.00E+00 & 0.00E+00 & + & 0.00E+00 & 0.00E+00 & + \\
 9 & -4.81E-01 & 3.34E-01 & \textbf{-5.11E-01} & \textbf{2.21E-01} & = & 0.00E+00 & 0.00E+00 & + & 0.00E+00 & 0.00E+00 & + & 0.00E+00 & 0.00E+00 & + & 0.00E+00 & 0.00E+00 & + \\
 10 & \textbf{-5.17E-01} & \textbf{3.96E-02} & -4.81E-01 & 1.44E-01 & + & 0.00E+00 & 0.00E+00 & + & 0.00E+00 & 0.00E+00 & + & 0.00E+00 & 0.00E+00 & + & 0.00E+00 & 0.00E+00 & + \\
 11 & \textbf{-8.70E-01} & \textbf{1.24E-01} & -3.55E-01 & 1.18E-01 & + & 0.00E+00 & 0.00E+00 & + & 0.00E+00 & 0.00E+00 & + & 0.00E+00 & 0.00E+00 & + & 0.00E+00 & 0.00E+00 & + \\
 12 & -1.18E-01 & 1.52E-07 & \textbf{-1.99E-01} & \textbf{2.19E-01} & - & 0.00E+00 & 0.00E+00 & + & 0.00E+00 & 0.00E+00 & + & 0.00E+00 & 0.00E+00 & + & 0.00E+00 & 0.00E+00 & + \\
 13 & -5.51E-01 & 4.49E-02 & \textbf{-6.16E-01} & \textbf{3.36E-01} & = & 0.00E+00 & 0.00E+00 & + & 0.00E+00 & 0.00E+00 & + & 1.24E-06 & 6.69E-06 & + & 3.48E-05 & 1.88E-04 & + \\
 14 & \textbf{-7.14E-01} & \textbf{1.48E-01} & 6.32E-01 & 3.97E+00 & + & 3.72E-03 & 8.23E-03 & + & 1.66E-03 & 5.23E-03 & + & 5.19E-03 & 9.40E-03 & + & 1.93E-01 & 4.74E-01 & + \\
 15 & \textbf{-3.47E+00} & \textbf{2.13E-02} & -3.45E+00 & 1.39E-01 & = & 0.00E+00 & 0.00E+00 & + & 0.00E+00 & 0.00E+00 & + & 0.00E+00 & 0.00E+00 & + & 0.00E+00 & 0.00E+00 & + \\
\hline
 W/T/L & \multicolumn{2}{c|}{$-/-/-$} & \multicolumn{3}{c|}{8/6/1} & \multicolumn{3}{c|}{15/0/0} & \multicolumn{3}{c|}{15/0/0} & \multicolumn{3}{c|}{15/0/0} & \multicolumn{3}{c|}{15/0/0} \\
\hline
\end{tabular}
\end{table}

\end{CJK}
\end{document}